\documentclass[5pt]{elsarticle}

\def\eg{e.g.}

\usepackage{multirow}
\usepackage{enumerate}
\usepackage{stfloats}
\usepackage{multicol}
\usepackage{algorithm}
\usepackage{amsmath}
\usepackage{amssymb}
\usepackage{amsthm}
\usepackage{color}
\usepackage{url}
\usepackage{subfigure}
\usepackage{graphics}
\newcommand{\cred}{\color{black}}

\begin{document}

\begin{frontmatter}



%\title{Dynamic Texture and Scene Classification: The Role of Spatial and Temporal Convolutional Neural Network}
%\title{Dynamic Texture and Scene Classification: The Role of Convolutional Statistical Features}
%\title{Dynamic Texture and Scene Classification: The Role of \\Statistical ConvNet Features}
\title{Statistical ConvNet Features for Dynamic Texture and Scene Classification}
\title{Dynamic texture and scene classification by transferring deep image features}

%\author[1]{Xianbiao Qi}
%\author[1]{Guoying Zhao}
%\author[1]{Matti Pietik{\"a}inen}

%
%
%
\author[1,2]{Xianbiao Qi}
\author[2]{Chun-Guang Li}
\author[1]{Guoying Zhao}
\author[1]{Xiaopeng Hong}
\author[1]{\\Matti Pietik{\"a}inen}
%%\cortext[cor1]{All correspondence should be addressed to Xianbiao Qi.}
%
%\address[1]{Department of Information and Telecommunication Engineering, Beijing University of Posts and %Telecommunications, Beijing 100876, China. E-mail: qixianbiao@gmail.com}
%\address[1]{Department of Computer Science and Engineering, Oulu university
\address[1]{Center for Machine Vision Research, University of Oulu, PO Box 4500, FIN-90014, Finland. E-mails: qixianbiao@gmail.com, \{gyzhao, xhong, mkp\}@ee.oulu.fi}
\address[2]{School of Information and Communication Engineering, Beijing University of Posts and Telecommunications, Beijing 100876, China. E-mail: lichunguang@bupt.edu.cn}

\begin{abstract}
Dynamic texture and scene classification are two fundamental problems in understanding natural video content. Extracting robust and effective features is a crucial step towards solving these problems. However the existing approaches
suffer from the sensitivity to either varying illumination, or viewpoint changing, or even camera motion, and/or the lack of spatial information. Inspired by the success of deep structures in image classification, we attempt to leverage a deep structure to extract feature for dynamic texture and scene classification. To tackle with the challenges in training a deep structure, we propose to transfer some prior knowledge from image domain to video domain. To be specific, we propose to apply a well-trained Convolutional Neural Network (ConvNet) as a mid-level feature extractor to extract features from each frame, and then form a representation of a video by concatenating the first and the second order statistics over the mid-level features. We term this two-level feature extraction scheme as a Transferred ConvNet Feature (TCoF). Moreover we explore two different implementations of the TCoF scheme, i.e., the \textit{spatial} TCoF and the \textit{temporal} TCoF, in which the mean-removed frames and the difference between two adjacent frames are used as the inputs of the ConvNet, respectively. We evaluate systematically the proposed spatial TCoF and the temporal TCoF schemes on three benchmark data sets, including DynTex, YUPENN, and Maryland, and demonstrate that the proposed approach yields superior performance.
\end{abstract}


\end{frontmatter}

\section{Introduction}
\label{sec:intro}
\begin{figure}
\begin{center}
 \includegraphics[width=1.0\linewidth]{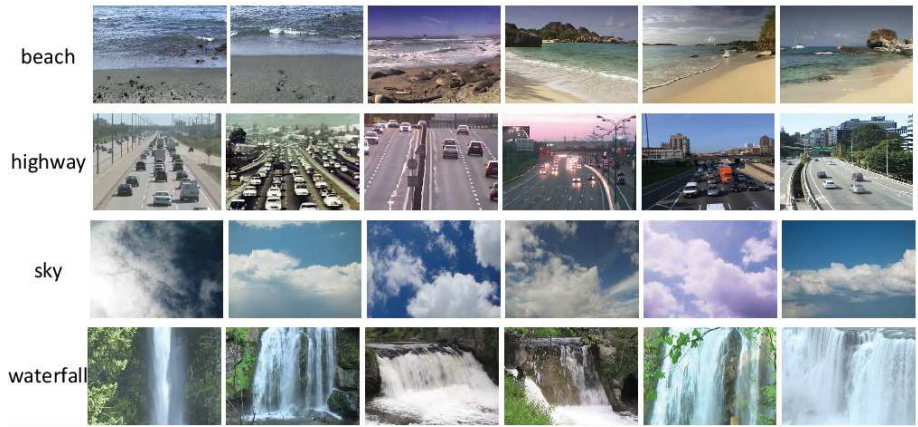}
\end{center}
   \caption{Sample images from dynamic scene data set YUPENN. Each row corresponds a category.} %In each row, each image comes from different sequences. It is easy to see that the sequences have high similarity in spatial appearance.}
\label{fig:sample-img-spatial-effectiveness}
\end{figure}

{\cred
Dynamic texture and dynamic scene classification are two fundamental problems in understanding natural video content and have gained considerable research attention~\cite{xu2011dynamic, peteri2010dyntex, doretto2003dynamic, zhao2007dynamic, derpanis2012spacetime, derpanis2010dynamic, ravichandran2013categorizing, chaudhry2013dynamic, theriault2013dynamic, feichtenhoferbags, derpanis2012dynamic, memiseviclearning, feichtenhofer2013spacetime, shroff2010moving}. Roughly, dynamic textures can be described as visual processes, which consist of a group of particles with random motions; dynamic scenes can be considered as places where events occur. In Fig.~\ref{fig:sample-img-spatial-effectiveness}, we show some sample images from a dynamic scene data set YUPENN \cite{derpanis2012dynamic}. The ability to automatically categories dynamic textures or scenes is useful, since it can be used to recognize the presence of events, surfaces, actions, and phenomena in a video surveillance system. %serve to provide priors for

However automatically categorizing dynamic textures or dynamic scenes is a challenging problem, since the existence of a wide range of naturally occurring variations in a short video, e.g., illumination variations, viewpoint changes, or even significant camera motions. %It is well known that constructing a robust and effective representation of a video sequence is the most critical component to approach these challenges. Consequently,
It is commonly accepted that constructing a robust and effective representation of a video sequence is a crucial step towards solving these problems. In the past decade, a large number of methods for video representation have been proposed, e.g., Linear Dynamic System (LDS) based methods~\cite{doretto2003dynamic, ravichandran2013categorizing, afsari2012group, chaudhry2013dynamic}, GIST based method~\cite{oliva2001modeling}, Local Binary Pattern (LBP) based methods~\cite{ojala2002multiresolution, zhao2007dynamic, pietikainen2011computer, qi2012pairwise, rahtu2012local}, and Wavelet based methods~\cite{derpanis2010dynamic, feichtenhoferbags, ji2013wavelet, xu2011dynamic}. Unfortunately, the existing approaches suffer from the sensitivity to either varying illumination, or viewpoint changing, or even the camera motion, and/or the lack of spatial information.}

Recently there is a surge of research interests in developing \textit{deep structures} for solving real world applications. Deep structure based approaches set up numerous recognition records in image classification~\cite{azizpour2014generic, sharif2014cnn}, object detection~\cite{sermanet2013overfeat}, face recognition and verification~\cite{Sun_2014_CVPR, sun2014deep}, speech recognition~\cite{deng2013recent}, and natural language processing~\cite{collobert2008unified, collobert2011natural}. %Among numerous deep structures, Convolutional Neural Networks (ConvNets) have been demonstrated to be extremely successful in computer vision~\cite{lecun1998gradient, krizhevsky2012imagenet, sermanet2013overfeat, jia2014caffe, chatfield2014return, azizpour2014generic}. % CNN based method has significantly flew many recognition records made by some carefully hand-crafted features in many applications, such as image classification, object detection, face detection, recognition and verification.
%\subsection{Our Contributions}
{\cred Inspired by the great success of deep structures in image classification, in this paper, we attempt to leverage a deep structure to extract feature for dynamic texture and scene classification. However, learning a deep structure needs huge amount of train data and is quite expensive in computational demand. Unfortunately, as in other video classification tasks, the dynamic textures and scenes classification tasks suffer from the small size of training data. As a result, the lack of training data is actually an obstacle to deploy a deep structure for video classification tasks.

By noticing of that there are a lots of work in learning deep structures for classifying images, in this paper, we attempt to transfer the knowledge in image domain to compensate the deficiency of training data in training a deep structure to represent dynamic textures and scenes.
%To tackle with the training challenges, we propose to transfer some prior knowledge from image domain. %the obstacle from the small size of training data and expensive challenge in computational demand
%\noindent {\bf Our Contributions.}
Concretely,
we propose to apply a well-trained Convolutional Neural Network (ConvNet) as a mid-level feature extractor to extract features from each frame in a video, and then form a representation of a video by concatenating the first and the second order statistics over the mid-level features. % for each video sequence.  is robust and effective.
We term this two-level feature extraction scheme as a Transferred ConvNet Feature (TCoF). %, In the proposed TCoF scheme, we apply a %of the output of a deep neural network structure
%In this paper, we propose a robust and effective feature extraction scheme, termed as Transferred ConvNet Features (TCoF), for representing dynamic textures and dynamic scenes.

Our aim in this paper is to explore a robust and effective way to capture the spatial and temporal information in dynamic textures and scenes. %Concretely, we propose to leverage the deep network to extract complex low-level features accommodate the ConvNets~\cite{krizhevsky2012imagenet} for extracting robust and effective representation of dynamic texture and dynamic scene.
To be specific, our contributions are highlighted as follows:
\begin{itemize}
\item %We propose a robust and effective feature extraction approach, termed as Statistical ConvNet Features (TCoF), for categorizing dynamic textures and dynamic scenes, %The proposed approach is inspired by the success of deep neural networks %Convolutional Neural Network (CNN) framework for static images categorization.
    We propose a two-level feature extraction scheme to represent dynamic textures and scenes, which applies a trained Convolutional Neural Network (ConvNet) as a feature extractor to extract mid-level features from each frame in a video and then computes the first and the second order statistics over the mid-level features. To the best of our knowledge, this is the first investigation of using a deep network with transferred knowledge to represent dynamic texture or scenes. %is used to extract mid-level features for categorizing dynamic textures and dynamic scenes.

    %For the first time, the deep network is used to extract mid-level features for categorizing dynamic textures and dynamic scenes. % to extract mid-level features and then constructs of mid-level features extracted by a trained ConvNets.
\item We investigate the effects of the spatial and temporal mid-level features %TCoF and the temporal TCoF by experiments
      on three benchmark data sets. Experimental results show that: a) the spatial feature is more effective for categorizing the dynamic textures and dynamic scenes and b) when the video is stabilized the temporal feature could provide some complementary information.
%\item We improve the state-of-the-art performance from 96.2\% to 99.1\% on data set YUPENN and from 76.2\% to 88.5\% on data set Maryland.
\end{itemize}

%
%Unlike images, representing a video sequence needs to consider the following aspects: % of information:
%\begin{enumerate}
%\item To depict the spatial information. In most cases, we can recognize the dynamic textures and scenes from a single frame in a video. Thus, extracting the spatial information of each frame in a video might be an effective way to represent the dynamic textures or scenes. %features by considering the video as multiple individual frames.
%\item To capture the temporal information. In dynamic textures or scenes, there are some specific micro-motion patterns. Capturing these micro-motion patterns might help to better understand the dynamic textures or scenes. % more clearly.
%\item To fuse the spatial and temporal information. When the spatial and temporal information are complementary,  combining both of them might boost the recognition performance.
%\end{enumerate}
%
%
%Different from the rigid or semi-rigid objects (e.g., actions), the dynamics of texture and scene are relatively random and non-directional. %In Fig.~\ref{fig:sample-img-spatial-effectiveness}, we show some sample images from a dynamic scene data set YUPENN \cite{derpanis2012dynamic}.
%Whereas the temporal information might be the most important cue in action recognition~\cite{marszalek2009actions}, it might be not that important, compared to the spatial information, in categorizing the dynamic textures or scenes. Nevertheless the temporal information might be complementary in some cases. %It is unclear in the literature by far.
}

%However the wide range of naturally occurring phenomena in the dynamic textures or scenes make categorizing and understanding them quite challengeable. It is commonly accepted that constructing robust and effective representation of a video sequence is a crucial step towards solving these problems.
%By far, the majority of studies focus on understanding 2D static images, e.g., which kind of texture is the image made of?  What is the scene in the image? It is well known that temporal information can greatly facilitate our understanding to the properties of the images.

The remainder of the paper is organized as follows. We review the related studies in Section~\ref{sec:related-work} % convolutional neural network in Section 2.
and present our proposals %the sptaial CNN and temporal CNN
in Section~\ref{sec:main-part}.  We evaluate the proposed spatial and temporal TCoF schemes % the S-CNN and T-CNN, and compare them with the state-of-the-art methods.
in Section~\ref{sec:experiments} and finally we conclude this paper with a discussion in Section~\ref{sec:conclusion}.

\section{Related Work}
\label{sec:related-work}

%\subsection{Dynamic Texture and Scene Classification}
%\label{sec:DT-and-DS}

In the literature, there are numerous approaches for dynamic texture and scene classification. While being closely relevant, dynamic texture classification \cite{xu2011dynamic, peteri2010dyntex, doretto2003dynamic, zhao2007dynamic, derpanis2012spacetime, derpanis2010dynamic, ravichandran2013categorizing, chaudhry2013dynamic} and dynamic scene classification \cite{theriault2013dynamic, feichtenhoferbags, derpanis2012dynamic, memiseviclearning, feichtenhofer2013spacetime, shroff2010moving} are usually considered separately as two different problems by far.
%\footnote{

{\cred
The research history of dynamic texture classification is much longer than that of the dynamic scene. The later, as far as we know, started since two dynamic scene data sets -- Maryland Dynamic Scene data set ``in the wild''~\cite{shroff2010moving} and York stabilized Dynamic Scene data set~\cite{peteri2010dyntex} -- were released. Although there might not be a clear distinction in nature, the slight difference of dynamic texture from dynamic scene is that the frames in a video of dynamic texture consist of images with richer texture whereas the frames in a video of dynamic scene are a natural scene involving over time. % motion information.
In addition, having mentioned of the data sets, compared to dynamic textures which are usually stabilized videos, the dynamic scene data set might include some significant camera motions.}

The critical challenges in categorizing the dynamic textures or scenes come from the wide range of variations around the naturally occurring phenomena. %It is commonly accepted that constructing robust and effective representation of a video sequence is a crucial step towards solving these problems.
%The difficulty to categorize dynamic textures or dynamic scenes arise from a wide range of naturally occurring variations: illumination variations, viewpoint changes, or even significant camera motions. %It is well known that constructing a robust and effective representation of a video sequence is the most critical component to approach these challenges. Consequently,
To overcome the difficulty, numerous methods for video representation have been proposed. % for dynamic textures or dynamic scenes.
Among them, Linear Dynamic System (LDS) based methods~\cite{doretto2003dynamic, ravichandran2013categorizing, afsari2012group, chaudhry2013dynamic}, GIST based method~\cite{oliva2001modeling}, Local Binary Pattern (LBP) based methods~\cite{ojala2002multiresolution, zhao2007dynamic, pietikainen2011computer, qi2012pairwise, rahtu2012local}, and wavelet based methods~\cite{derpanis2010dynamic, feichtenhoferbags, ji2013wavelet, xu2011dynamic} are the most widely used. LDS is a statistical generative model which captures the spatial appearance and dynamics in a video~\cite{doretto2003dynamic}. While LDS yields promising performance on viewpoint-invariant sequences, it performs poor on viewpoint-variant sequences~\cite{ravichandran2013categorizing, afsari2012group, chaudhry2013dynamic}. Besides, it is also sensitive to illumination variations. %Different from LDS, LBP based methods are geometrically driven approach.
GIST~\cite{oliva2001modeling} represents the spatial envelope of an image (or a frame in video) holistically by Gabor filter. However GIST suffers from scale and rotation variations.
Among LBP based methods, Local Binary Pattern on Three Orthogonal Planes (LBP-TOP)~\cite{zhao2007dynamic} is the most widely used. LBP-TOP describes a video by computing local binary pattern from three orthogonal planes ($xy$, $xt$ and $yt$) \textit{only}. After LBP-TOP, several variants have been proposed, e.g., Local Ternary Pattern on Three Orthogonal Planes (LTP-TOP) \cite{rahtu2012local}, Weber Local Descriptor on Three Orthogonal Planes (WLD-TOP)~\cite{chen2013automatic}, Local Phase Quantization on Three Orthogonal Planes (LQP-TOP)~\cite{rahtu2012local}. While LBP-TOP and its variants are effective at capturing spatial and temporal information and robust to illumination variations, they are suffering from camera motions.
Recently, wavelet based methods are also proposed, e.g., Spatiotemporal Oriented Energy (SOE)~\cite{feichtenhoferbags}, Wavelet Domain Multifractal Analysis (WDMA)~\cite{ji2013wavelet}, and Bag-of-Spacetime-Energy (BoSE)\cite{feichtenhoferbags}. Combined with the Improved Fisher Vector (IFV) encoding strategy~\cite{perronnin2010improving, chatfield2011devil}, BoSE leads to the state-of-the-art performance on dynamic scene classification. %The performance of BoSE comes from the advanced encoding methods. Meanwhile,
However, the computational cost of BoSE is expensive due to slow feature extraction and quantization.

The aforementioned methods can be roughly divided into two categories: the \textit{global} approaches and the \textit{local} approaches.
The \textit{global} approaches extract features from each frame in a video sequence by treating each frame as a whole, e.g., LDS~\cite{doretto2003dynamic} and GIST~\cite{oliva2001modeling}. %in GIST, color information in each channel is used individually.
While the global approaches describe the spatial layout information well, they suffer from the sensitivity to illumination variations, viewpoint changes, or scale and rotation variations.
The \textit{local} approaches construct a statistics (e.g., histogram) on a bunch of features extracted from local patches in each frame or local volumes in a video sequence, %Majority of the existing methods belong to this category,
including LBP-TOP~\cite{zhao2007dynamic}, LQP-TOP~\cite{rahtu2012local}, BoSE~\cite{feichtenhoferbags}, Bag of LDS~\cite{ravichandran2013categorizing}. While the local approaches are robustness to transformations (e.g., rotation, illumination), they suffer from the lack of spatial layout information which is important to represent a dynamic texture or dynamic scene.

{\cred
%\noindent {\bf Our Motivations.} In sprite of the successes of deep structures in image classifications, we propose to %it is a natural idea to
%deploy a deep structure for classifying dynamic textures and scenes. However, learning a deep structure needs huge amount of train data, and is quite expensive in computational demand. Unfortunately, as in other video classification tasks, the dynamic textures and scenes classification tasks suffer from the small size of training data. %While \textit{deep structures} based approaches have been shown a lots of successes, to learn a deep structure. %neural networks,
%As a result, the lack of training data is actually an obstacle to deploy a deep structure for all video classification tasks. %\footnotte{In this paper, we consider only the dynamic textures and scenes classification. Our approach is also applicable to other suitable tasks.}.

%By noticing that there are a lots of work in learning deep structures for classifying images, i

In this paper, we attempt to leverage a deep structure with transferred knowledge from image domain to construct a robust and effective %to compensate the deficiency of training data in order to apply a deep structure for classifying
representation for dynamic textures and scenes. To be specific, we propose to use a pre-trained ConvNet -- which has been trained on the large-scale image data set ImageNet~\cite{krizhevsky2012imagenet}, \cite{sharif2014cnn}, \cite{azizpour2014generic} -- as transferred (prior) knowledge, %which can be served as an initialization,
and then fine-tune the ConvNet with the frames in the videos of training set. % when necessary.
Equipped with a trained ConvNet, we extract mid-level features from each frame in a video and represent a video by the concatenation of the first and the second order statistics over the mid-level features. %We term our approach to represent a video sequence as Transferred ConvNet Features (TCoF).
% Transferred ConvNet Statistical Features (TCoF).
%Statistical (Transferred??) ConvNet Features (STCoF).
% In this paper, we attempt to represent the dynamic textures and dynamic scenes via Convolutional Neural Network (ConvNet) based statistical features.

%To the best of our knowledge, this is the first investigation of using a deep network with transferred knowledge to represent dynamic texture or scenes. %is used to extract mid-level features for categorizing dynamic textures and dynamic scenes.
Compare to previous studies, our approach possesses the following advantages:
\begin{itemize}

% it is as a general feature extractor.
\item Our approach represents a video with a two-level strategy. The deep structure used in the frame level is easier to train or even train-free, since we can adopt prior knowledge from image domain.
    %The adopted initialization, as a transferred knowledge from the large-scale image data set ImageNet, provides remarkable representation power. In addition, by using a good initialization, it is much fast and easier to train well a ConvNet than with random initialization.

\item The extracted frame-level features are robust to translations, small scale variations, partial rotations, and illumination variations.

\item Our approach represents a video sequence by a concatenation of the first and the second order statistics of the frame-level features. %, in which each frame in the video contributes a ConvNet feature vector.
    This process is fast and effective.

\end{itemize}

}
In the next section, we will present the framework and two different implementations of our proposal.

\section{Our Proposal: Transferred ConvNet Feature (TCoF)}
\label{sec:main-part}
% \label{sec:spatial-TCoF}

%Dynamic texture and scene can be seen as texture and scene motion along with the time. In one side, human can recognition the texture and scene properties from a single still image, in another side, temporal information can enhance human's understanding to the texture and scene. Thus, how to capture these two kinds of information effectively is an extremely important problem in dynamic texture and scene classification. In this section, we will analysis the dynamic texture and scene from spatial and temporal viewpoints independently.

%\subsection{TCoF} % Spatial TCoF, Statistical ConvNet Feature

%In this section, we present our Statistical ConvNet Features (TCoFs).
Our TCoF scheme consists of three stages:
\begin{itemize}
\item Constructing a ConvNet with transferred knowledge from image domain; %a good initialization; % by using a pre-trained ConvNet model~\cite{krizhevsky2012imagenet}
\item Extracting the mid-level feature with the ConvNet from each frame in a video;
\item Forming the video-level representation by concatenating the calculated first and the second order statistics over the frame-level features.
\end{itemize}

%Given a trained ConvNet, we construct a TCoF representation of a video sequence %The scheme of extracting TCoF is illustrated
%as the flowchart illustrated in Fig.~\ref{fig:STCNN}.

%\subsection{Renaissance of Convolutional Neural Network for Image Classification}
\subsection{Convolutional Neural Network with Transferred Knowledge for Extracting Frame-Level Features}

\begin{figure}
\begin{center}
 \includegraphics[width=0.85\linewidth]{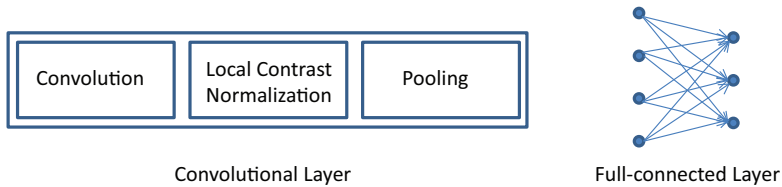}
\end{center}
   \caption{The typical structure of a ConvNet. }
\label{fig:cnn}
\end{figure}

Notice that there are a lots of work in learning deep structures for classifying images. Among them, Convolutional Neural Networks (ConvNets) have been demonstrated to be extremely successful in computer vision~\cite{lecun1998gradient, krizhevsky2012imagenet, sermanet2013overfeat, jia2014caffe, chatfield2014return, azizpour2014generic}. % CNN based method has significantly flew many recognition records made by some carefully hand-crafted features in many applications, such as image classification, object detection, face detection, recognition and verification.

We show a typical structure of a ConvNet in Fig.~\ref{fig:cnn}. The ConvNet consists of two types of layers: convolutional layers and full-connected layers. The convolutional part, as shown in the left panel of Fig.~\ref{fig:cnn}, consists of three components -- convolutions, Local Contrast Normalization (LCN), and pooling. Among the three components, the convolution block is compulsory, %necessary, prerequisite,
and LCN and the pooling are optional. The convolution components capture complex image structures. The LCN achieves invariance to illumination variations. The pooling component can \textit{not only} yield partial invariance to scale variations and translations, but \textit{also} reduce the complexity for the downstream layers. Due to sharing parameters which is motivated by the \textit{local reception field} in biological vision system, the number of free parameters in the convolutional layer are significantly reduced. The full-connected layer, as shown in the right panel of Fig.~\ref{fig:cnn}, is the same as a multi-layer perception neural network. %In the ConvNets framework, the full-connected layers have the most parameters needed to learn from data.

\begin{figure*}[t]
\begin{center}
 \includegraphics[width=1.0\linewidth]{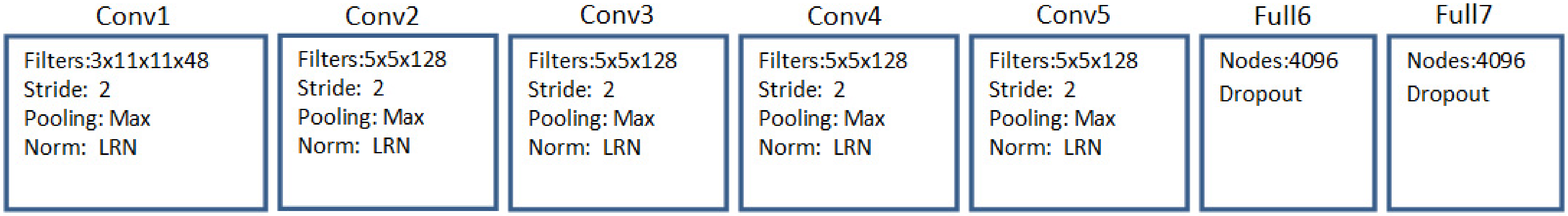}
\end{center}
   \caption{The architecture of the ConvNet used in our TCoF scheme.}%~\cite{krizhevsky2012imagenet}
\label{fig:ConvNet-5-2}
\end{figure*}

%Notice that there are a lots of work in learning deep structures for classifying images. Among them,

%The most successful ConvNet implementation is introduced by Krizhevsky et al.~\cite{krizhevsky2012imagenet}, which won the large-scale ImageNet contest. %Their ConvNet consists of five convolutional layers and three full-connected layers (in which the last layer is the classification layer), and is robust to sorts of image transformations.\footnote{For example, max-pooling tactics makes ConvNets robust to translations, small scale variations, and partial rotations, and LCN makes ConvNets robust to illumination variations.}
{\cred
In our TCoF framework, we use a ConvNet with five convolutional layers and two full-connected layers as shown in Fig.~\ref{fig:ConvNet-5-2}, which is the same as the most successful ConvNet implementation introduced by Krizhevsky et al.~\cite{krizhevsky2012imagenet} and won the large-scale ImageNet contest, to extract the mid-level feature from each frame in a video. Note that we remove the final full-connected layer in the ConvNet introduced in~\cite{krizhevsky2012imagenet}.

As mentioned previously, training well a deep network like that in Fig.\ref{fig:ConvNet-5-2} needs huge mount of training data and is quite expensive in computational demand. In our case, for dynamic texture or scene, the training data is limited. % it is an obstacle to train a huge network with a few
%To train the ConvNet,
In stead of training a deep network from scratch, which is quite time-consuming, we propose to use the pre-trained ConvNet~\cite{krizhevsky2012imagenet} %~\cite{sharif2014cnn},~\cite{azizpour2014generic},which has been trained on the large-scale data set ImageNet,
as the initialization, and fine-tune the ConvNet with the frames in videos from training data if necessary. %\footnote{}.
By using a good initialization, we virtually transfer miscellaneous prior knowledge from image domain (e.g., data set ImageNet) to the dynamic textures and scenes tasks.

%The success of ConvNet in the large-scale ImageNet contest greatly stimulates the research interests. Lots of effort focuses on extending ConvNet to different applications. For example, Razavian et al.~\cite{sharif2014cnn} and Azizpour et al.~\cite{azizpour2014generic} demonstrate that ConvNets can be generalized to object recognition, fine-grained classification, action recognition, and etc.
%Note that we take the weights in each layers but the final full connection layer in the well-trained ConvNet~\cite{krizhevsky2012imagenet} as the initialization. The final layer can be initialized in a supervised way separately, in which the output of the first seven layers served as input feature. % and the train the single final layer .
%And then, the whole ConvNet is fine-tuned with the train data. % and in our feature extraction scheme. However, the pre-trained other layers are kept.}
%This training process is fast owning to a good initialization. The fine-tuning is much easier than training a ConvNet from scratch with random initialization.

}
%The ConvNet extracts a ConvNet feature from each frame in a video sequence.

\subsection{Construct Video-Level Representation}
\label{sec:video-level-representation}

%Then we form TCoF by concatenating the calculated statistics. The scheme of extracting TCoF is illustrated in Fig.~\ref{fig:STCNN}. In our TCoF framework, the used ConvNet consists of five convolutional layers and two full-connected layers.\footnote{What should be pointed out is that we do not keep the final classifier layer of the original ConvNet~\cite{krizhevsky2012imagenet} in our feature extraction framework. However, the pre-trained other layers are kept.} The ConvNet extracts a ConvNet feature from each frame in a video sequence.
% However in training stage, the classifier layer was kept~\cite{krizhevsky2012imagenet}.

Given a video sequence containing $N$ frames, the ConvNet yields $N$ ConvNet features. Note that as the input to the ConvNet in TCoF, we use each frame in a video subtracting \textit{an averaged image}. % which is precalculated on the whole ImageNet data set~\cite{krizhevsky2012imagenet}.

%\begin{equation}
%F_{S-CNN} = FC(I(i) - mI), \\\\\
%\label{eq:SCNN}
%\end{equation}
%where I(i) means the $i$th frame of the input sequence, and mI denotes the mean image obtained by averaging the whole ImageNet data set. For FC(.) from each frame, we normalize it with $L_2$ norm.

Denote $X$ as a set of the ConvNet features $\{\textbf{x}_1, \textbf{x}_2, ..., \textbf{x}_N\}$ where $\textbf{x}_i \in R^d$ is the ConvNet feature extracted from $i$-th frame. % of dimension $d$. %Here, we regard the sequence features as a set. Given a set,
We extract the \textit{first} and the \textit{second order} statistics on feature set $X$.

The first-order statistics of $X$ is the \textit{mean} vector which is defined as follows:
\begin{equation}
\textbf{u} = \frac{1}{N} \sum_{i=1}^{N} \textbf{x}_i,
\label{eq:mean}
\end{equation}
where $\textbf{u}$ captures the average behaviors of the $N$ ConvNet features which reflect the average characteristics in the video sequence. %The first-order can be considered as the mean of feature in each dimension. It reflects the mean response of CNN features on all frames.

The second-order statistics is the \textit{covariance} matrix which is defined as follows:
\begin{equation}
\textbf{S} = \frac{1}{N} \sum_{i=1}^{N}(\textbf{x}_i-\textbf{u})(\textbf{x}_i-\textbf{u})^\top,
%\textbf{Cov} = \frac{1}{N} \sum_{i=1}^{N}\sum_{j=1}^{N} (\textbf{x}_i-\textbf{u})(\textbf{x}_j-\textbf{u})^{T},
\label{eq:cov}
\end{equation}
where $\textbf{S}$ describes the variation of the $N$ ConvNet features from the mean vector $\textbf{u}$ and the correlations among different dimensions.
%the covariance matrix is usually considered as an effective second-order statistics. The covariance matrix captures the correlation between the data in any two dimension.
The dimension of covariance feature is $\frac{d\times (d+1)}{2}$. % since its symmetry property. %, thus, in practice, only the upper triangular matrix is used. Thus, t.
When $d$ is large (e.g., $d = 4096$), the dimension of the covariance feature is high.
%Due to the high dimension of the second-order features, we assume that the between-in dimensions are independent. Under this assumption,
Instead, we propose to extract \textit{only} the diagonal entries in $\textbf{S}$ as the second-order feature, that is,
\begin{equation}
\textbf{v} = \text{diag}(\textbf{S}),
\label{eq:var}
\end{equation}
where $\text{diag}(\cdot)$ means to extract the diagonal entries of a matrix as a vector. The vector $\textbf{v}$ is $d$-dimensional and captures the variations along each dimension in the ConvNet features. %characteristic fluctuations over times  features which reflect the average behaviors in the video sequence.
%, the final dimension for our second-order feature is 4096.

Having calculated the first and the second order statistics, we form the video-level representation, TCoF, by concatenating $\textbf{u}$ and $\textbf{v}$, i.e.,
\begin{equation}
\textbf{f} =\left[
  \begin{array}{c}
    \textbf{u} \\
    \textbf{v} \\
  \end{array}
\right],
\label{eq:TCoF}
\end{equation}
where the dimension of a TCoF representation is $2d$.

\begin{figure*}
\begin{center}
 \includegraphics[width=1.0\linewidth]{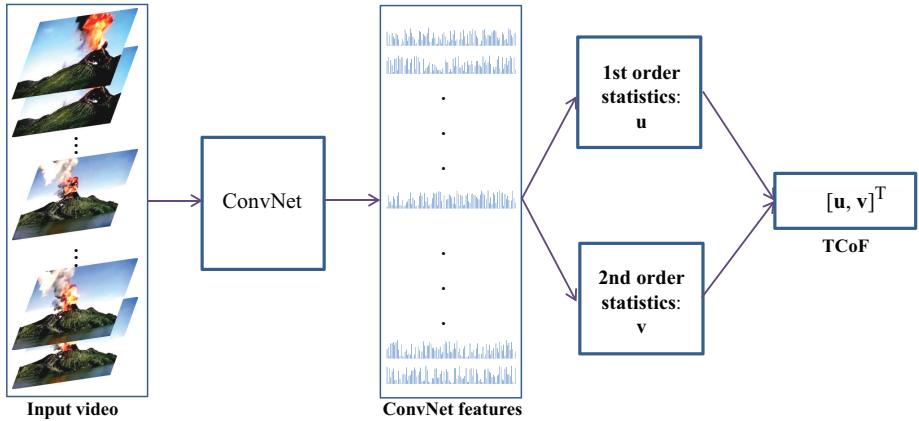}
\end{center}
%   \caption{An illustration of Spatial and Temporal Convolutional Neural Network.}
   \caption{An illustration of our TCoF scheme.} % the scheme of extracting
\label{fig:STCNN}
\end{figure*}

%Given a trained ConvNet,
For clarity, we illustrate the flowchart of constructing a TCoF representation for a video sequence in Fig.~\ref{fig:STCNN}.

\noindent {\bf Remarks 1.}
Our proposed TCoF belongs to global approach. Since the spatial layout information can be captured well, we term the TCoF scheme described above as the \textit{spatial} TCoF. Our proposed TCoF possesses the robustness to translations, small scale variations, partial rotations, and illumination variations owing to the ConvNet component. In addition, the process of extracting a TCoF vector is extremely fast since that the ConvNet adopt a so-called \textit{stride} tactics and the second step in TCoF is to calculate the two statistics.

\subsection{Modeling Temporal Information}
\label{sec:temporal-TCoF}

%Dynamic textures and scenes are an extension to the texture and scene to the temporal domain.
While it is well accepted that dynamic information can enrich our understanding of the textures or scenes, modeling the dynamic information is difficult. % is much harder to depict than spatial information.
Unlike the motion of rigid object, dynamic texture and scene are usually involving of non-rigid objects and thus the optical flow information seems relatively random. % more incorrect.

In this paper, we propose to use the difference of the adjacent two frames in a short-time to capture the random-like micro-motion patterns. %is an effective way to depict the motion.
To be specific, we take the difference between the $(i+\tau)$-th and $i$-th frames as the input of the ConvNet component in TCoF scheme, where $\tau \in \{1, \cdots, N-1\}$ is an integer which corresponds to the resolution in time to capture the random-like micro-motion patterns. In practice, we set $\tau$ as a small integer, \eg, 1, 2 or 3.

Given a video sequence containing $N$ frames, the ConvNet produces $N-\tau$ \textit{temporal} frame-level features. Then we extract the \textit{first} and the \textit{second order} statistics on the temporal ConvNet features to form a temporal TCoF for the input video, in the same way as the spatial TCoF in Section~\ref{sec:video-level-representation}.

%\begin{equation}
%F_{T-CNN} = FC(I(t_{i+\tau} )-I(t_i))), \\\\\
%\label{eq:TCNN}
%\end{equation}
%where $I(t_{i} )$ is the $i$th frame of the sequence, and  $I(t_{i+\tau} )$ is the $i+\tau$th frame. FC(.) denotes a 4096-D feature. For FC(.) from each frame, we normalize it with $L_2$ norm.

\noindent {\bf Remarks 2.}
The temporal TCoF differs from the spatial TCoF in the input of the ConvNet. In the spatial TCoF, we take each frame in a video subtracting a precalculated average image as input; whereas in the temporal TCoF we take the difference of two frames in a short-time and %unlike the spatial TCoF,
there is no need to subtract an average image. % in the temporal TCoF.

\noindent {\bf Remarks 3.}
In our proposed TCoF, we treat the extracted $N$ ConvNet features as a set and ignore %. Note that by treating the $N$ features as a set,
the sequential information among features. % is discarded.
The rationale of this simplification comes from the property of dynamic textures and dynamic scenes. Note that the dynamic textures are visual processes of a group of particles with random motions, and dynamic scenes are places where natural events are occurring, the sequential information %micro-motions
in these processes are relatively random and thus %the sequential information is no longer that
less critical. Experimental results in Section~\ref{sec:evaluation-spatial-temporal-TCoF} support this point.

\section{Experiments}
\label{sec:experiments}

%\color{blue}
In this section, we introduce the benchmark data sets, the baseline methods, and the implementation details, and then present the experimental evaluations of our approach. % on three benchmark data sets.

\subsection{Data Sets Description}
\begin{figure}
\begin{center}
 \includegraphics[width=0.95\linewidth]{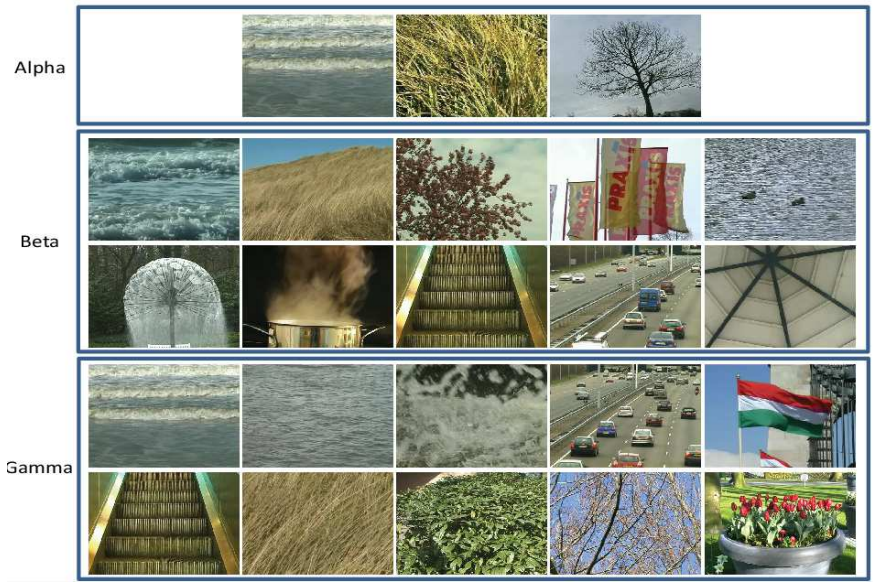}
\end{center}
   \caption{Sample frames from DynTex data set. The ``Alpha'' , ``Beta'', and ``Gamma'' show the sample frames from each category in the data set.} %The ``Gamma'' shows five videos from the same category ``Fountain''. The ``Gamma'' set contains complex image variations, e.g., scale, orientation.}
\label{fig:dynSamples}
\end{figure}

{\bf{DynTex}}~\cite{peteri2010dyntex} %\footnote{http://projects.cwi.nl/dyntex/}
is a widely used dynamic texture data set, containing 656 videos with each sequence recorded in PAL format. % ($720 \times 576$, 25 fps).
The sequences in DynTex are divided into three data subsets -- ``Alpha'',
``Beta'' and ``Gamma'': %Some detailed information about these three data sets is listed below:
%\begin{itemize}
%\item
a) ``Alpha'' data subset contains 60 sequences which are equally divided into 3 categories: ``sea'', ``grass'' and ``trees'';
%\item
b) ``Beta'' data subset consists 162 sequences which are grouped into 10 categories: ``sea'', ``grass'', ``trees'', ``flags'', ``calm water'', ``fountains'', ``smoke'', ``escalator'', ``traffic'', and ``rotation''; %Compared to ``Alpha'' data set, more complex phenomena are added into this data set.
%\item
c) ``Gamma'' data subset is composed of 264 sequences which are grouped into 10 categories: ``flowers", ``sea'', ``trees without foliage'', ``dense foliage'', ``escalator'', ``calm water'', ``flags'', ``grass'', ``traffic'' and ``fountains''. Compared to ``Alpha'' and ``Beta'' data subsets, this data subset contains more complex image variations, e.g., scale, orientation, and etc.
%Sample images from the ``Alpha'', ``Beta'' and ``Gamma'' are shown in Figure \ref{fig:dynSamples}
%\end{itemize}
Sample frames from the three data subsets are shown in Fig.~\ref{fig:dynSamples}.

\begin{figure}
\begin{center}
 \includegraphics[width=1.0\linewidth]{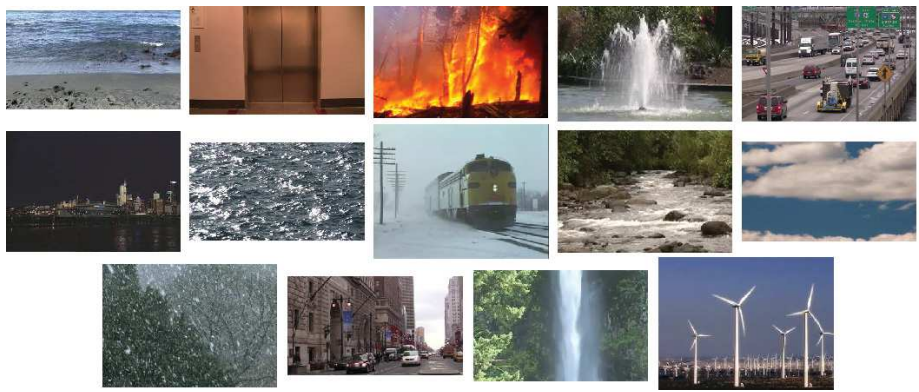}
\end{center}
   \caption{Samples from dynamic scene data set YUPENN. Each image corresponds to a category of video sequence.} %For(left-to-right, top-to-bottom) beach, city street, elevator, forest fire, fountain, highway, lighting room, ocean, railway, rushing river, sky-clouds, snowing, waterfull and windmill farm.}
\label{fig:yupenn}
\end{figure}

{\bf{YUPENN}} \cite{derpanis2012dynamic} %\footnote{www.cse.yorku.ca/vision/research/dynamic-scenes.shtml}
is a ``stabilized'' dynamic scenes data set. This data set was introduced to emphasize scene-specific temporal information. YUPENN consists of fourteen dynamic scene categories with 30 color videos in each category.
%The average resolution of the videos are $250 \times 370$ with 145 frames.
%The videos are obtained from different sources. Due to the diversity within and across the video sources,
The sequences in YUPENN have significant variations, such as frame rate, scene appearance, scale, illumination, and camera viewpoint. Some sample frames are shown in Fig.~\ref{fig:yupenn}.

\begin{figure}
\begin{center}
 \includegraphics[width=1.0\linewidth]{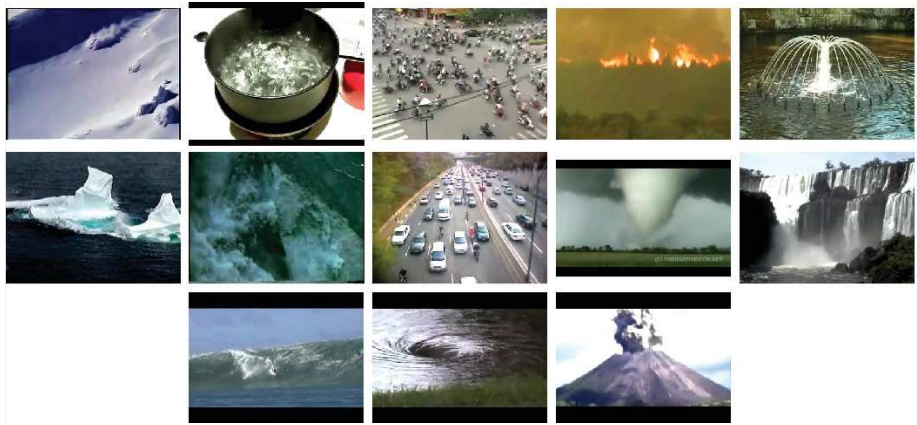}
\end{center}
   \caption{Sample frames from Maryland scenes data set. } % (left-to-right, top-to-bottom) Avalanche, Boiling water, Chaotic Traffic, Forest Fire, Fountain, Iceberg Collapse, Landslide, Smooth Traffic, Tornado, Volcanic Eruption, Waterfall, Waves and Whirlpool.}
\label{fig:maryland}
\end{figure}

{\bf{Maryland}} \cite{shroff2010moving} %\footnote{http://www.umiacs.umd.edu/~nshroff/DynamicScene.html}
is a dynamic scene data set which was introduced firstly. It consists of 13 categories with 10 videos per category. %These videos were downloaded from video hosting website, like ``Youtube''. As there is no control over the video capturing process,
The data set have large variations in illumination, frame rate, viewpoint, and scale. Besides, there are variations in resolution and camera dynamics. %These variations ensure that the intra-class variation is very high. This data set is harder than the YUPENN data set.
Some sample frames are shown in Fig.~\ref{fig:maryland}. % shows some sample frames from each category.

\subsection{Baselines and Implementation Details}
\label{baselines-implementations}

{\bf{Baselines}} We compare our proposed TCoF approach with the following state-of-the-art methods\footnote{For the LBP-TOP, we report the results with our own implementation and for other methods we cite the results from their papers.}.
\begin{itemize}
\item GIST \cite{oliva2001modeling}: Holistic representation of the spatial envelope which is widely used in 2D static scene classification.

\item Histogram of Flow (HOF) \cite{marszalek2009actions}: The HOF is an well-known descriptor in action recognition. % It is considered to always perform better than its spatial counterpart Histogram of Orientation Gradient (HOG).

\item Local Binary Pattern on Three Orthogonal Planes (LBP-TOP) \cite{zhao2007dynamic}: The LBP-TOP is widely used in dynamic texture, dynamic facial expression, and dynamic facial micro-expression. %, and considered as a baseline for these applications.

\item Chaotic Dynamic Features (Chaos) \cite{shroff2010moving}. %The Chaos was proposed to understand the dynamic scene system via chaos systems.

\item Slow Feature Analysis (SFA) \cite{theriault2013dynamic}. % The SFA is a principle that can be used to learn local motion descriptor in an unsupervised way.
%\item[6) ]
\item Synchrony Autoencoder (SAE) \cite{memiseviclearning}.
%\item[7) ]
\item Synchrony K-means (SK-means) \cite{memiseviclearning}.
%\item[8) ]
\item Complementary Spacetime Orientation (CSO) \cite{feichtenhofer2013spacetime}: In CSO, the complementary spatial and temporal features are fused in a random forest framework.
%\item[9) ]
\item Bag of Spacetime Energy (BoSE) \cite{feichtenhoferbags}.  %BoSE tried to improve the dynamic scene recognition from three aspects: (i) feature extraction, (ii) feature encoding, and (iii) the pooling strategy. They used the spacetime orientation energy feature, improved fisher vector encoding and adaptive pooling strategy of the encoded features.
\end{itemize}

{\bf{Implementation Details.}} In both the spatial TCoF (s-TCoF) and the temporal TCoF (t-TCoF), we resize the frame into $224\times 224$ and normalize both s-TCoF and t-TCoF with $L_2$-norm, respectively. For the combination of both the spatial and temporal TCoF, we take the concatenation of the two normalized s-TCoF and t-TCoF and denote it as st-TCoF. We do not use any data augmentation method. %, although it has been proved that data augmentation may improve the performance.
For our t-TCoF, we use $\tau = 3$. We use the CAFFE toolbox~\cite{jia2014caffe} to extract the proposed TCoFs.
%To train the ConvNet,
%We use the pre-trained ConvNet~\cite{krizhevsky2012imagenet} %,~\cite{sharif2014cnn},~\cite{azizpour2014generic} which has been trained on the large-scale data set ImageNet
%as an initialization. %, and fine-tune the ConvNet with the frames in videos from training data set.
Note that we take the weights in each layers but the final full connection layer in the well-trained ConvNet~\cite{krizhevsky2012imagenet} as the initialization. %The final layer is initialized in a supervised way separately, in which the output of the first seven layers served as input feature. % and the train the single final layer .
And then, the whole ConvNet could be fine-tuned with the train data. % and in our feature extraction scheme. However, the pre-trained other layers are kept.} This training process is fast owning to a good initialization.
While the fine-tuning stage is easier than training a ConvNet from scratch with random initialization, we observed that the improvement by the extra fine-tuning was minor. Thus we use the ConvNet without a further fine-tuning to extract the mid-level features.\footnote{Note that we use the Leave-One-Out cross-validation to evaluate the performance. The training data are changed from each trial. If we chose to fine-tune the ConvNet, we should fine-tune for each trial. Since the improvements were minor, we report the results without a fine-tuning to keep all experimental results are repeatable.}
For LBP-TOP, we use the best performing setting of $\text{LBP-TOP}_{8,8,8,1,1,1}$, and  % which is better than other configurations in our experiments.
the ${\chi}^2$ kernel. % for LBP-TOP feature.
To fairly compare with previous methods, we test our approach and other baselines with both the nearest neighbor (NN) classifier and SVM classifier separately. In SVM, we use a linear SVM with Libsvm toolbox \cite{chang2011libsvm}, in which the tradeoff parameter $C$ is fixed to 40 in all our experiments. Following the standard protocol, %configurations,
we use Leave-One-Out (LOO) cross-validation. %protocol. % is used for all experiments. For instance, in Maryland, they have 130 videos. For each time, one video is left for test, the other 129 videos are used for train. The reader can use the code\footnote{link.......................} to reproduce the result and figures reported in paper.
%The following experiments consist of two parts, dynamic texture classification and dynamic scene classification.
% of the First and Second Order Statistics, and

\subsection{Evaluation of the Spatial and Temporal TCoFs}
\label{sec:evaluation-spatial-temporal-TCoF}
In this subsection, we evaluate systematically the influence of %using first-order and second-order statistics,
using the spatial and temporal TCoFs on DynTex, YUPENN, and Maryland data sets.
%For convenience, in the following evaluation, we only use 1st order Spatial or Temporal TCoF features.
%, since they are captured in a ``stabilized'' condition.

%{\bf{Influence of the first and second order statistics in TCoF.}}
%We conduct a set of experiments to evaluate the performance of using different types of statistics to define our proposed TCoF. Experimental results are shown in Table~\ref{table:Orderfeatures}. % breakdown
%
%We observe from Table~\ref{table:Orderfeatures} that .....
%the s-TCoF only using one image has performed well on both tasks, and almost saturates the performance.

{\bf{Effectiveness of the $s$-TCoF.}}
Since that the s-TCoF features are constructed by accumulating all features in all frames, it is interesting to investigate the effect to the final performance of using different number of frames. To this end, we evaluate the s-TCoF using seven different settings: 1) using only the \textit{first} frame in a video, 2) using the first $\frac{N}{8}$ frames in a video, 3) using the first $\frac{N}{4}$ frames in a video, 4) using the first $\frac{N}{2}$ frames in a video, and 5) using all $N$ frames in a video.
%For convenience, we only use the first-order statistics and show
Experimental results are shown in Table~\ref{table:spatialfeatures}.

\begin{table}[ht]                                 % used for centering table
\caption{Evaluation of the spatial TCoF (s-TCoF) by using different number of frames. }              % title of Table
%\scriptsize
\small
\centering
\begin{tabular}{c c c c c c}          % centered columns (4 columns)
\hline
% \footnote{only one image is used.}
Datasets  & 1st  & $\frac{N}{8}$  & $\frac{N}{4}$ & $\frac{N}{2}$ & $N$\\
\hline
Alpha(NN)        &100     & 100     & 100    & 100    & 100   \\

Beta (NN)        &98.77   & 99.38   & 99.38  & 98.77  & 99.38   \\

Gamma(NN)        &97.73   & 97.35   & 96.97  & 96.97  & 96.59   \\
\hline
YUPENN(NN)       &95.71   & 96.43   & 96.19  & 96.43  & 95.48   \\

YUPENN(SVM)       &96.90   & 96.90   & 96.90  & \textbf{97.14} & 96.90   \\ %\textbf{97.14}
\hline

Maryland(NN)       &72.31   & \textbf{80.00}   & 75.38  & 77.69  & 76.92   \\
Maryland(SVM)       &80.00   & 83.85   & 80.77  & 83.08  & \textbf{88.46}   \\

\hline
\end{tabular}
\label{table:spatialfeatures}                   % is used to refer this table in the text
\end{table}

From Table \ref{table:spatialfeatures}, we can see that the spatial TCoF performs well by even using the first frame \textit{only}, on DynTex and YUPENN data sets. %However on data set Maryland, all frames a
This confirmed the effectiveness of the spatial TCoF scheme.
%Or on both tasks, and almost saturates the performance.
%This observation has validated that the spatial information is discriminative in DTC and DSC.

\begin{table}[ht]
\caption{ Evaluating the performance of temporal TCoF (t-TCoF) as a function of parameter $\tau$. }              % title of Table
\centering                                     % used for centering table
\small
\begin{tabular}{c c c c c c }          % centered columns (4 columns)
\hline
Datasets  & $\tau = 1$ & $\tau = 2$ &  $\tau = 3$ &  $\tau = 4$ &  $\tau = 5$\\
\hline
Alpha(NN)        & 98.33  &  96.67  & 96.67  & 96.67 & 96.67\\
Beta (NN)        & 97.53  &  96.91  & 97.53  & 97.53 & 97.53\\
Gamma(NN)        & 93.56  &  94.32  & 93.18  & 93.18 & 93.94\\
\hline
YUPENN(NN)       & 90.24  &  91.19  & 92.38  & 93.57 & 93.10\\
YUPENN(SVM)       & 94.52  &  96.19  & 96.67  & 96.90 & 97.14\\
\hline
Maryland(NN)       & 55.38  &  56.92  & 57.69  & 61.54 & 63.85\\
Maryland(SVM)       & 66.92  &  62.31  & 61.54  & 63.85 & 63.85\\
\hline
\end{tabular}
\label{table:temporalfeatures}                   % is used to refer this table in the text
\end{table}

{\bf{Effectiveness of the $t$-TCoF.}} Here, we conduct experiments to evaluate the influence of the parameter $\tau$. % to the performance. %For convenience, we only use the first-order features, and
Experimental results are shown in Table \ref{table:temporalfeatures}. We observe from Table~\ref{table:temporalfeatures} that t-TCoF is not sensitive to the choice of $\tau$. %, the t-TCoF works well. This partially validates the effectiveness of the temporal feature.

{\bf{Comparison of s-TCof and t-TCoF.}}
Note that almost all the results of s-TCoF in Table~\ref{table:spatialfeatures} outperform that of t-TCoF in Table~\ref{table:temporalfeatures}. This suggest that s-TCoF is more effective than t-TCoF, since the randomness of the micro-motions in dynamic texture or natural dynamic scene makes the temporal information less critical. Nevertheless t-TCoF could provide complementary information to s-TCoF in some case that will be shown later.

\subsection{Comparisons with the State-of-the-Art Methods} % Dynamic Texture Classification
%According to the above-mentioned evaluation of the effectiveness of temporal features, we will use $\tau = 3$ for all data sets.

%\subsubsection{Dynamic Texture Classification}
{\bf{Dynamic Texture Classification on DynTex Data Set}}
We conduct a set of experiments to compare our methods with LBP-TOP. %, s-TCoF, t-TCoF and st-TCoF. Overall classification accuracy
Experimental results are shown in Table \ref{table:dyntexresults}.
%and a detail classification confusion table is shown in Figure \ref{fig:confusiondyntex}.

\begin{table}[ht]
\caption{Classification Results on DynTex dynamic texture data set. The performance of LBP-TOP is based on our implementation. All methods use NN classifier.}              % title of Table
\centering                                     % used for centering table
\small
\begin{tabular}{c c c c c}          % centered columns (4 columns)
\hline
Datasets    & LBP-TOP  &  s-TCoF  &  t-TCoF  &st-TCoF\\
\hline
Alpha        & 96.67   & \textbf{100}      & 96.67   & 98.33\\
%\hline
Beta         & 85.80   & \textbf{99.38}    & 97.53   & 98.15\\
%\hline
Gamma        & 84.85   & 95.83    & 93.56   & \textbf{98.11}\\
\hline
\end{tabular}
\label{table:dyntexresults}                   % is used to refer this table in the text
\end{table}

We observe from Table \ref{table:dyntexresults} that:
\begin{enumerate}
\item The s-TCoF performs the best on data subsets Alpha and Beta. This results confirm that the s-TCoF is effective for dynamic texture classification.
%\item It is easy to find that with the increase of the difficulties of the data sets, the performance of the LBP-TOP decreases a lot. For ``Alpha'' to ``Gamma'', the performance decreases from 96.67\% to 84.85\%. Compared to LBP-TOP, the t-TCoF is robust.
\item On Gamma subset, s-TCoF and t-TCoF significantly outperform LBP-TOP. Moreover by combining s-TCoF with t-TCoF, we achieve the best result. This result suggests that t-TCoF might provide complementary information to s-TCoF.
%\item The s-TCoF and t-yTCoF features are complementary. On ``Gamma'' set, the st-TCoF obtains  that outperforms both s-TCoF and t-TCoF.

\end{enumerate}

{\bf{Dynamic Scene Classification on YUPENN. }}
We compare our methods with the state-of-the-art methods, including CSO, GIST, SFA, SOE, SAE, BOSE, SK-means, and LBP-TOP, and the experimental results are presented in Table \ref{table:yupennresults} and %. The category-wise accuracy for all methods are shown in
Table \ref{table:yupenndetail}. We observe from Table \ref{table:yupennresults} that:
\begin{enumerate}
\item The s-TCoF and t-TCoF both outperform the state-of-the-art methods. %s except the combination of s-TCoF and t-TCoF. This suggests that the spatial information is extremely important for scene understanding. GIST is also a way to capture spatial information, but my s-TCoF greatly outperforms it. This explain that CNN is an effective way to represent the scene images.
    Note that YUPENN consist of stabilized videos. These results confirm that both s-TCoF and t-TCoF are effective for dynamic scene data in a stabilized setting.

\item The combination of the s-TCoF and t-TCoF, i.e., the st-TCoF, performs the best, in which reduce the error relatively over $30\%$. As shown is Table~\ref{table:yupenndetail}, s-TCoF and t-TCoF are complementary to each other on some categories, e.g., ``Light Storm'', ``Railway'', ``Snowing'', and ``Wind. Farm''.

     %The st-TCoF significantly outperforms other methods, such as s-TCoF, t-TCoF and BoSE. The st-TCoF with 1-NN achieves 98.57\% that has significantly improves the previous best performance reported in \cite{feichtenhoferbags} (96.19\%). This may suggest that, in ``stabilized'' dynamic scenes, the s-TCoF and t-TCoF are complementary. As shown in Table \ref{table:yupennresults}, the st-TCoF outperforms both of the s-TCoF and t-TCoF.

%\item The performance of SVM consistently outperforms that of NN. This suggests suggest that better classifiers can be designed to improve the performance further.

\end{enumerate}

\begin{table*}[ht]
\caption{Classification results on scene data set YUPENN. The results of are taken from the corresponding papers. The performance of LBP-TOP is based on our implementation.}              % title of Table
\centering                                     % used for centering table
%\small
\scriptsize
\begin{tabular}{@{\ }c@{\ } @{\ }c@{\ } @{\ }c@{\ } @{\ }c@{\ } @{\ }c@{\ } @{\ }c@{\ } @{\ }c@{\ } @{\ }c@{\ } @{\ }c@{\ } @{\ }c@{\ } @{\ }c@{\ } @{\ }c@{\ } }  %@{\ }c@{\ } @{\ }c@{\ }        % centered columns (4 columns)
\hline
  Methods   & CSO  & GIST &  SFA  &  SAE &  SOE &  BoSE &  SK-means &  LBP-TOP & s-TCoF   & t-TCoF  &  st-TCoF \\
\hline
NN          & -    & 56   & -     & 80.7   & 74      & -       & -  & 75.95   & 96.43    & 93.10 & \bf{98.81}  \\
%\hline
SVM         & 85.95& -    & 85.48 & 96.0    & 80.71   & 96.19   & 95.2  & 84.29 & 97.14   & 97.86 & \bf{99.05}
\\
\hline
\end{tabular}
\label{table:yupennresults}                   % is used to refer this table in the text
\end{table*}

\begin{table*}[ht]
\caption{Category-wise accuracy ($\%$) for different methods on dynamic scene data set YUPENN. All methods use linear SVM classifier. Our methods and LBP-TOP are based on our implementation. The other results are taken from \cite{feichtenhoferbags}.}              % title of Table
\centering                                     % used for centering table
\scriptsize
%\small
\begin{tabular}{@{\ }c@{\ } @{\ }c@{\ }  @{\ }c@{\ }  @{\ }c@{\ }  @{\ }c@{\ }  @{\ }c@{\ }  @{\ }c@{\ }  @{\ }c@{\ }  @{\ }c@{\ }  @{\ }c@{\ }  @{\ }c@{\ } }
\hline
  \multirow{2}{*}{Categories}  & HOF+  & Chaos+    &   \multirow{2}{*}{SOE}   & \multirow{2}{*}{SFA}    & \multirow{2}{*}{CSO}   & \multirow{2}{*}{LBP-TOP} & \multirow{2}{*}{BoSE}  & \multirow{2}{*}{s-TCoF}  & \multirow{2}{*}{t-TCoF}  &  \multirow{2}{*}{st-TCoF}\\
         & GIST      & GIST          &         &        &       &       &    &    &        &        \\
\hline
Beach           & 87        & 30            &  93     & 93     & 100  & 87& 100   & 97    & 97    & 97    \\

Elevator        & 87        & 47            &  100     & 97     & 100 & 97 & 97   & 100    & 100    & 100    \\

Forest Fire     & 63        & 17            &  67     & 70     & 83   & 87 & 93   & 100    & 97    & 100    \\

Fountain        & 43        & 3            &  43     & 57     & 47    & 37& 87   & 100    & 97    & 100    \\

Highway         & 47        & 23            &  70     & 93     & 73   & 77 & 100   & 97    & 100    & 100    \\

Light Storm  & 63        & 37            &  77     & 87     & 93   & 93& 97   & 90    & 100    & 100    \\

Ocean           & 97        & 43            &  100     & 100     & 90 & 97 & 100   & 100    & 100    & 100    \\

Railway         & 83        & 7            &  80     & 93     & 93    &80 & 100   & 97    & 100    & 100    \\

Rush River   & 77        & 10            &  93     & 87     & 97   & 100 & 97   & 97    & 97    & 97    \\

Sky-Clouds      & 87        & 47            &  83     & 93     & 100  &93 & 97   & 100    & 97    & 100  \\
Snowing         & 47        & 10            &  87     & 70     & 57   &83 & 97   & 90    & 97    & 97    \\

Street          & 77        & 17            &  90     & 97     & 97   &93 & 100   & 100    & 97    & 100    \\
Waterfall       & 47        & 10            &  63     & 73     & 77   &90 & 83   & 93    & 93    & 97    \\

Wind. Farm   & 53        & 17            &  83     & 87     & 93   & 67& 100   & 100    & 100    & 100    \\
\hline
Overall         & 68.33     & 22.86         &  80.71  & 85.48  & 85.95& 84.29&96.19   & 97.14    & 97.86    & \bf{99.05}    \\
\hline
\end{tabular}
\label{table:yupenndetail}                   % is used to refer this table in the text
\end{table*}

%\noindent
{\bf{Dynamic Scene Classification on dynamic scene data set Maryland. }}
%In this experiments, we also study the effect of better classifiers on the recognition performance of dynamic scenes. We perform the comparative experiments between the nearest neighbor (NN) a linear SVM. Overall accuracies for relevant methods are
We present the comparison of our methods with the state-of-the-art methods in Table \ref{table:marylandresults}. % and %. We also give a category-wise accuracy
%in Table \ref{table:marylandCategory}.
We observe from Table \ref{table:marylandresults} that:
\begin{enumerate}

\item The s-TCoF significantly outperforms the other methods. This suggests that the spatial information is extremely important for scene understanding. %GIST is also a way to capture spatial information, but my s-TCoF greatly outperforms it. This explain that CNN is an effective way to represent the scene images.

\item The results of t-TCoF are much worse than s-TCoF. This might be due to the significant camera motions in this data set.

%\item The performance of SVM consistently outperforms that of NN. For instance, s-TCoF with a linear SVM improves the performance from 76.92\% to 88.46\%. This suggests suggest that better classifiers can be designed to improve the performance further.

\end{enumerate}

\begin{table*}[ht]
\caption{Classification results on scene data set Maryland. The results of are taken from the corresponding papers. The performance of LBP-TOP is based on our implementation.}              % title of Table
\centering                                     % used for centering table
%\small
\scriptsize
\begin{tabular}{@{\ }c@{\ \ }  @{\ \ }c@{\ \ } @{\ \ }c@{\ \ } @{\ \ }c@{\ \ } @{\ \ }c@{\ \ } @{\ \ }c@{\ \ } @{\ \ }c@{\ \ } @{\ \ }c@{\ \ } @{\ \ }c@{\ \ } @{\ \ }c@{\ }}         % centered columns (4 columns)
\hline
  Methods   & CSO  &  SFA   &  SOE &  BoSE  &  LBP-TOP &  s-TCoF &  t-TCoF &  st-TCoF\\
\hline
NN          & -    & -      & -    & -     & 31.54   & 74.62  & 58.46   & 74.62           \\
%\hline
SVM         & 67.69& 60     & 43.1 & 77.69 & 39.23   & \bf{88.46}  & 66.15   & \bf{88.46}           \\
\hline
\end{tabular}
\label{table:marylandresults}                   % is used to refer this table in the text
\end{table*}

\subsection{Further Investigations and Remarks} % Remarks
\label{sec:further-investigation-discussion}

\begin{figure*}
\begin{center}
 \includegraphics[width=1.0\linewidth]{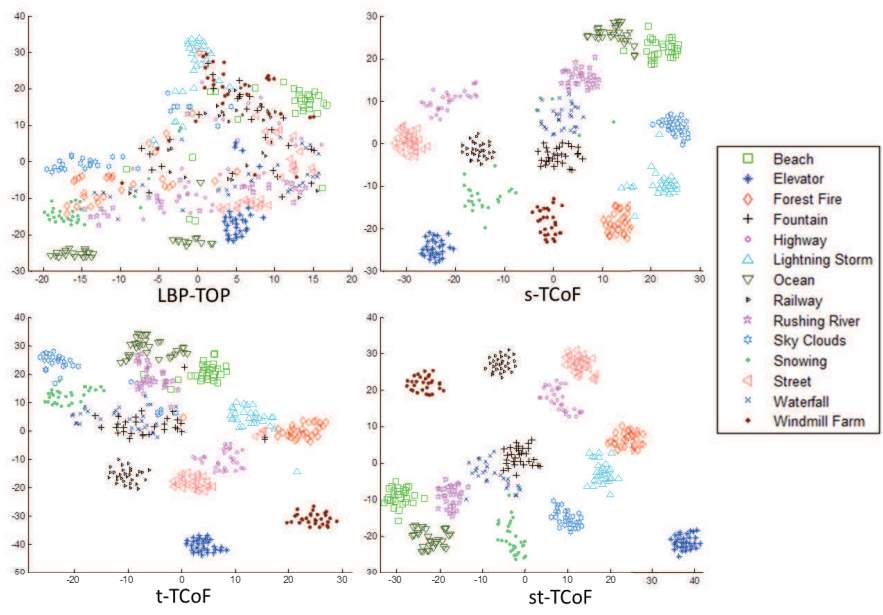}
\end{center}
   \caption{Data visualization of LBP-TOP, s-TCoF, t-TCoF, and st-TCoF on YUPENN dynamic scene data set. Each point in the figure corresponds to a video sequence.}
\label{fig:yorkSNE}
\end{figure*}

\begin{figure*}
\begin{center}
 \includegraphics[width=0.95\linewidth]{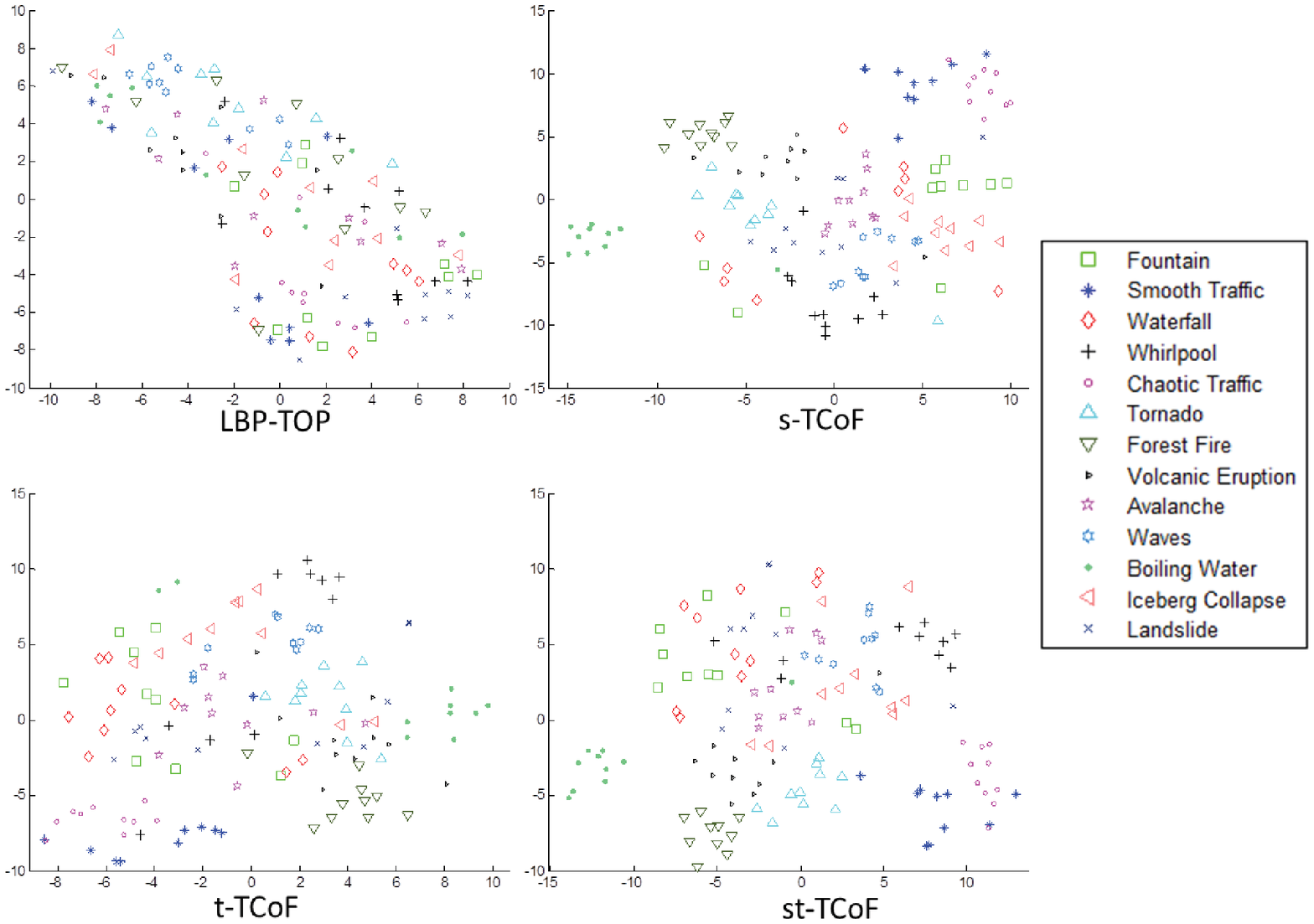}
\end{center}
   \caption{Data visualization of LBP-TOP, s-TCoF, t-TCoF, and st-TCoF on Maryland dynamic scene data set. Each point in the figure corresponds to a video sequence.}
\label{fig:marylandSNE}
\end{figure*}

{\bf{Data Visualization. }} To show the discriminative power of the proposed approach, we use
t-Stochastic Neighbor Embedding (t-SNE)\footnote{t-SNE is a (prize-winning) technique for dimensionality reduction that is particularly well suited for the visualization of high-dimensional data.} \cite{van2008visualizing} to visualize the data distributions of the dynamic scene data sets YUPENN and Maryland.
Results are shown in Fig.~\ref{fig:yorkSNE} and Fig.~\ref{fig:marylandSNE}, respectively. We observe from Fig.~\ref{fig:yorkSNE} and Fig.~\ref{fig:marylandSNE} that s-TCoF, t-TCoF, and st-TCoF yield distinct separations between categories. These results reveal the effectiveness of our proposed TCoF approach vividly.

% Among them, st-TCoF

{\bf{Remarks. }}
Note that in our TCoF schemes, we treat the frames in a video as orderless images and extract mid-level features with a ConvNet. By doing so, the sequential information among features is ignored. The superior experimental results suggest that such a simplification is harmless. %The rationale of this simplification comes from the property of dynamic textures and dynamic scenes.
The sequential information in these processes contributes less (or not at all) discriminativeness, because that the dynamic textures can be viewed as visual processes of a group of particles with random motions, and the dynamic scenes are places where natural events are occurring. %, the sequential information is no longer that critical.
The effectiveness underlying our proposed TCoF approach for dynamic texture and scene classification is due to the following aspects:
\begin{itemize}
\item Rich filters' combination built on color channels in ConvNet describes richer structures and color information. The filters that are built on different image patches capture stronger and richer structures compared to the hand-crafted features. % (e.g., SIFT, LBP).

\item ConvNet makes the extracted features robust to sorts of image transformations due to the max-pooling and LCN components. Specifically the max-pooling tactics makes ConvNet robust to translations, small scale variations, and partial rotations, and LCN makes ConvNet robust to illumination variations.

\item The first and the second order statistics %extracted by s-TCoF and t-TCoF
capture enough %the key
information over the mid-level features. % of the ConvNet features effectively.

\item When the video sequences are stabilized, the t-TCoF might provide complementary information to the s-TCoF.

\end{itemize}

%The spatial information  captured by s-TCoF is more important to dynamic textures or dynamic scenes classification. Nevertheless when video sequences are stabilized, the temporal information captured by t-TCoF might provide complementary information to s-TCoF.

\section{Conclusion and Discussion}
\label{sec:conclusion}

We have proposed a robust and effective feature extraction approach for dynamic texture and scene classification, termed as Transferred ConvNet Features (TCoF), which was built on the first and the second order statistics of the mid-level features extracted by a ConvNet with transferred knowledge from image domain. %deep neural network -- ConvNet --
%on video sequence. %TCoF, we compute the first-order and second-order statistics of the output of a pre-trained deep neural network -- ConvNet -- for video sequence.
We have investigated two different implementations of the TCoF scheme, i.e., the \textit{spatial} TCoF and the \textit{temporal} TCoF. We have evaluated systematically the proposed approaches on three benchmark data sets and confirmed that: a) the proposed \textit{spatial} TCoF was effective, and b) the \textit{temporal} TCoF could provides complementary information when the camera is stabilized. %Extensive experiments have demonstrated that the proposed approach could yield superior performance.

{\cred
Unlike images, representing a video sequence needs to consider the following aspects: % of information:
\begin{enumerate}
\item To depict the spatial information. In most cases, we can recognize the dynamic textures and scenes from a single frame in a video. Thus, extracting the spatial information of each frame in a video might be an effective way to represent the dynamic textures or scenes. %features by considering the video as multiple individual frames.
\item To capture the temporal information. In dynamic textures or scenes, there are some specific micro-motion patterns. Capturing these micro-motion patterns might help to better understand the dynamic textures or scenes. % more clearly.
\item To fuse the spatial and temporal information. When the spatial and temporal information are complementary,  combining both of them might boost the recognition performance.
\end{enumerate}

Different from the rigid or semi-rigid objects (e.g., actions), the dynamics of texture and scene are relatively random and non-directional. %In Fig.~\ref{fig:sample-img-spatial-effectiveness}, we show some sample images from a dynamic scene data set YUPENN \cite{derpanis2012dynamic}.
Whereas the temporal information might be the most important cue in action recognition~\cite{marszalek2009actions, karpathy2014large}, %it might be not that important, compared to the spatial information, in categorizing the dynamic textures or scenes. Nevertheless the temporal information might be complementary in some cases. %It is unclear in the literature by far.
%Dynamic textures are visual processes of a group of particles with random motions, and dynamic scenes are places where natural events are occurring. Compare to action recognition, the motions in dynamic textures or dynamic scenes are relatively random. Our
our investigation in this paper suggests that the sequential information in dynamic textures or scenes is \textit{not that critical} for classification. %Therefore we conclude that: a) the spatial information captured by s-TCoF is more important to classify dynamic textures or dynamic scenes; nevertheless, b) when video sequences are stabilized and the micro-motions in the dynamic process is less random, the temporal information captured by t-TCoF might provide complementary information to s-TCoF.
%The proposed TCoF approach is robust, efficient, and effective. Owning to these merits, we believe it is competent to be applied in \textit{Big Data} scenario.
}

\section*{Acknowledgement}
X. Qi, G. Zhao, X. Hong, and M. Pietik{\"a}inen are supported in part by the Academic of Finland and InfoTech.
%This research was supported in part by the Academic of Finland.
C.-G. Li is supported partially by the Scientific Research Foundation for the Returned Overseas Chinese Scholars, the Ministry of Education, China.
The authors would like to thank %Prof.
Renaud P\'{e}teri, %Prof.
Richard P. Wildes and  %Prof.
Pavan Turaga for sharing the DynTex dynamic texture, YUPENN dynamic scene, and Maryland dynamic scene data sets.

\bibliographystyle{elsarticle-harv}
\small
%\bibliography{<egbib>}
\bibliography{egbib}

\end{document}